\documentclass[runningheads]{llncs}

\usepackage{graphicx}
\usepackage{amsmath,amssymb}
\usepackage{hyperref}
\usepackage{booktabs}
\usepackage{multirow}
\usepackage{xcolor}
\usepackage{microtype}
\usepackage[most]{tcolorbox}
\usepackage{subcaption}
\usepackage{float}
\usepackage{xurl}
\usepackage{algorithm}
\usepackage{algorithmic}
\usepackage{caption}
\graphicspath{{figures/}}

\def\ours{\texttt{ByDeWay-V2}}

\begin{document}

\title{Explainable and Resource-Efficient Spatial Reasoning in Multimodal LLMs for Decision-Critical Applications}

\titlerunning{Explainable Spatial Reasoning in MLLMs}

\author{Piyush Jain\inst{1} \and  Kousik Dasgupta\inst{2} \and Rajarshi Roy\inst{3} \and Subarna Tripathi\inst{4}}

\institute{Heritage Institute of Technology, Kolkata, India \\
\email{piyushj0204@gmail.com} \and
Kalyani Government Engineering College, Kalyani, India \\
\email{kousik.dasgupta@gmail.com} \and  IAIRO, India \\\email{rajarshi.roy@iairo.ai} \and
Intel Corporation, USA \\
\email{subarna.tripathi@intel.com}}

\maketitle

\begin{abstract}
As Multimodal Large Language Models (MLLMs) are increasingly deployed in
decision-critical pipelines---robotics, embodied AI, and safety monitoring---the
opacity of their spatial judgments limits operator trust and auditability.
MLLMs demonstrate impressive reasoning capabilities but often struggle with
fine-grained spatial understanding and suffer from object hallucination.
Previous work, ByDeWay, introduced
Layered-Depth-Based Prompting (LDP), a training-free framework that
mitigates hallucinations by structuring input prompts using monocular
depth estimations.
However, while coarse depth layering effectively anchors depth
perception, it falls short in resolving complex object-to-object
spatial relationships within the same geometric plane---such as
projective (e.g., ``left of'', ``above'') and topological
(e.g., ``inside'', ``touching'') relations.
In this paper, we propose \textbf{ByDeWay-V2}, an
enhancement
that
seamlessly integrates explicit spatial relational context alongside
depth cues, expressed as human-readable predicates that serve as
auditable evidence for downstream decision support.
Leveraging an open-vocabulary object detector
(YOLO-World-L), our framework computes
precise pairwise geometric relations between detected objects and
injects them as structured spatial predicates into the MLLM prompt.
This fully training-free approach bridges the gap between 3D scene
depth and 2D spatial semantics.
We evaluate ByDeWay-V2 on the Visual Spatial Reasoning
(VSR) and BLINK benchmarks
across multiple MLLMs, with hallucination grounding assessed on POPE.
On the BLINK spatial subset, ByDeWay-V2 achieves a \textbf{46\%
relative improvement in F1 score} over LDP for Qwen2.5-VL, and
recovers BLIP-Base's spatial reasoning on VSR from near-random
performance to a competitive F1 of 0.53.
Notably, our lightest configuration operates under a strict 40-token
context budget on CPU, illustrating the framework's suitability for
resource-constrained, real-time decision-support settings.
\end{abstract}

\keywords{
Multimodal Large Language Models
\and Spatial Reasoning
\and Explainable AI
\and Decision Support Systems
\and Hallucination Reduction by Spatial Grounding
\and Resource-Efficient Inference
}

\section{Introduction}

Multimodal Large Language Models (MLLMs) have emerged as powerful
general-purpose systems capable of handling diverse tasks spanning both
language and vision modalities.
From visual question answering (VQA) to complex scene understanding,
these models exhibit strong performance in zero-shot and few-shot
settings by leveraging large-scale pretrained representations.
However, two persistent challenges continue to limit their reliability.
Chief among them are \textit{visual hallucinations}---the tendency of
MLLMs to generate content not grounded in the visual
input~\cite{bai2024hallucination}---and \textit{spatial reasoning
deficits}, where models struggle to accurately interpret geometric,
projective, and topological relationships between objects.
These issues become especially problematic in domains requiring precise
object-level grounding, such as robotics, embodied AI, and
safety-critical applications.

Beyond raw accuracy, deployment in decision-critical pipelines demands
\textit{interpretability}: an operator relying on an MLLM's spatial
judgment---e.g., to trigger a robotic arm's motion or flag a
safety violation---needs auditable justification for that judgment,
not merely a black-box output.
Existing training-free hallucination-mitigation methods largely
improve accuracy without producing intermediate, human-readable
evidence, limiting their usefulness in settings where decisions must
be explained or reviewed.
This motivates a framework that not only improves spatial reasoning
accuracy but also surfaces its reasoning as explicit, inspectable
predicates suitable for decision-support and auditing workflows.

\textbf{The spatial reasoning gap:}
While LDP provides crucial depth-aware structure, it operates at a
macro-regional level.
A significant limitation arises when querying fine-grained interactions
\textit{between} objects that co-exist within the same depth stratum.
Knowing that a ``person'' and a ``refrigerator'' both reside in the
``closest'' layer does not explicitly inform the MLLM whether the
person is ``next to'', ``in front of'', or ``inside'' the
refrigerator.
Benchmarks such as VSR~\cite{liu2023vsr} and BLINK~\cite{fu2024blink}
expose this weakness: modern MLLMs still frequently guess or
hallucinate answers when confronted with adjacency, orientation, or
topological queries.
We empirically confirm this gap---LDP alone yields an F1 score of only
0.05 for BLIP-Base on VSR, and as low as 0.50 on the BLINK spatial
reasoning subset for a state-of-the-art model such as Qwen2.5-VL.

\begin{figure}[H]
  \centering
  \includegraphics[width=\textwidth]{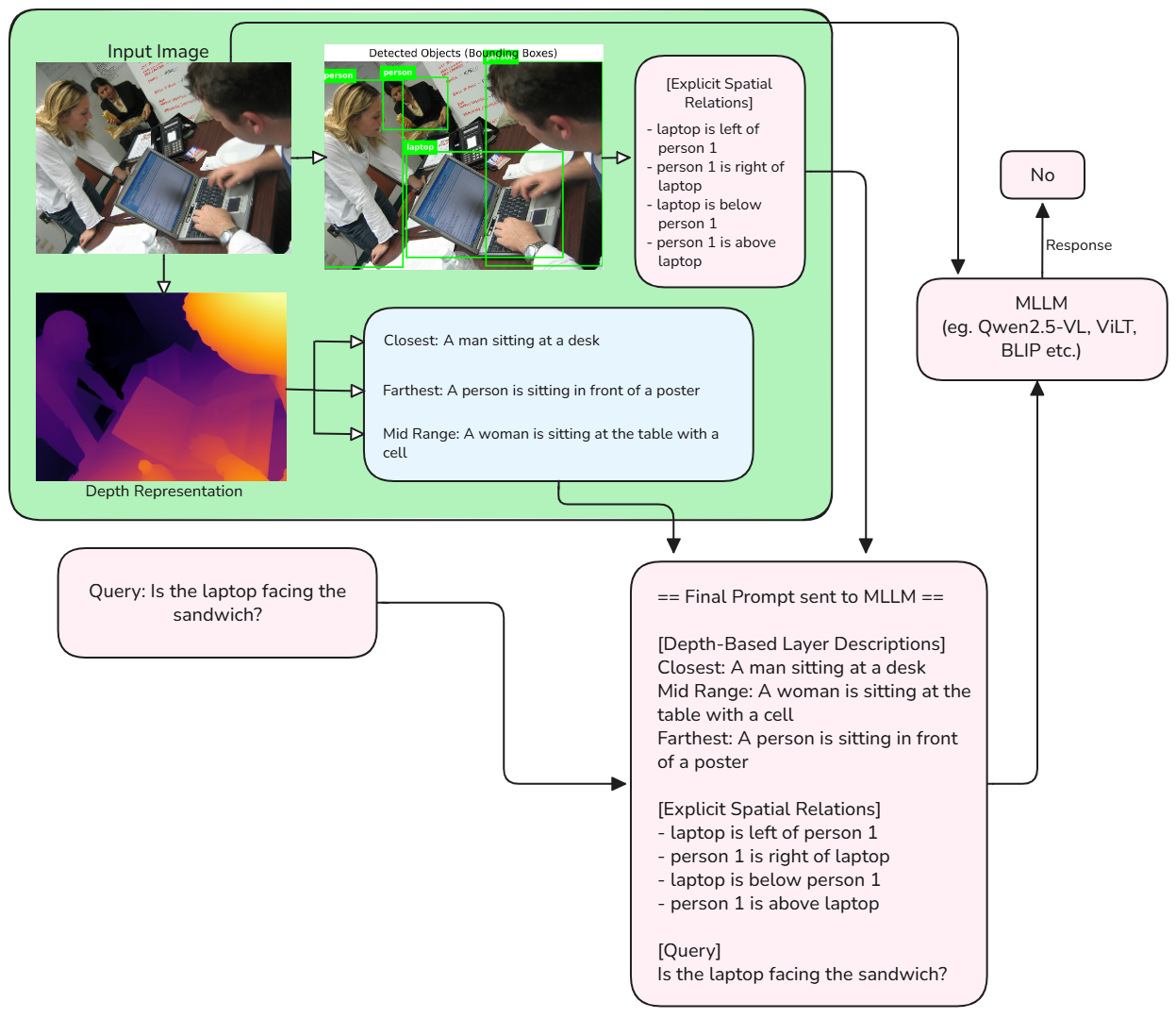}
  \caption{%
    \textbf{ByDeWay-V2 Workflow.}
    The framework runs two parallel branches on the input image.
    \textit{(Top branch)} YOLO-World-L detects all objects and
    returns precise bounding boxes; pairwise geometric heuristics
    then derive explicit spatial predicates
    (e.g.\ ``laptop is left of person~1'',
    ``person~1 is above laptop'').
    \textit{(Bottom branch)} DepthAnything~V2 produces a dense depth
    map that is partitioned into three layers (closest, mid-range,
    farthest); KOSMOS-2 captions each layer independently.
    Both streams are serialised into a single structured prompt
    (\textit{Final Prompt sent to MLLM}) and forwarded---together
    with the original image and user query---to the downstream MLLM
    (e.g.\ Qwen2.5-VL, ViLT, BLIP).
    The dotted path represents the standard baseline prompt used
    for comparison.
  }
  \label{fig:pipeline}
\end{figure}

To overcome this limitation, we present \textbf{ByDeWay-V2}.
We augment the original LDP pipeline with a dedicated
\texttt{SpatialAnalyser} module that extracts explicit geometric
spatial relations between detected objects.
We employ YOLO-World-L~\cite{cheng2024yoloworld}, an open-vocabulary
real-time detector, to identify and localise all relevant objects in
the scene via precise bounding boxes.
By computing pairwise geometric heuristics---including
Intersection-over-Union (IoU), centre-point distances, and bounding
box overlap ratios---we systematically map spatial interactions into
explicit, human-readable predicates (e.g., ``The laptop is left of
person 1'', ``The cat is inside the vase'').
This fine-grained relational context is appended alongside the
depth-layer captions, providing the MLLM with a comprehensive prompt
that marries 3D depth organisation with 2D inter-object spatial
semantics.
The complete pipeline is illustrated in Figure~\ref{fig:pipeline}.

As illustrated in Figure~\ref{fig:pipeline}, the ByDeWay-V2 pipeline operates through two parallel branches to process the input image. The upper branch employs YOLO-World-L to detect and localise objects, applying geometric heuristics to generate pairwise spatial predicates. Concurrently, the lower branch utilises DepthAnything~V2 and KOSMOS-2 to segment the scene into depth-specific text captions. These outputs are subsequently synthesised into a unified, structured prompt that equips the MLLM with comprehensive spatial and depth awareness without requiring parameter updates.
Our key contributions in ByDeWay-V2 are:
\begin{itemize}
    \item We introduce an enhanced, fully training-free framework that
    augments LDP with a \texttt{SpatialAnalyser} module for explicit
    spatial relation extraction.

    \item We present a hybrid prompting methodology that combines
    coarse depth-aware scene decomposition with precise,
    bounding-box-derived pairwise geometric predicates.
    \item We demonstrate the modular, model-agnostic nature of our
    approach, requiring no parameter updates to any downstream MLLM.

    \item We conduct extensive evaluations on VSR~\cite{liu2023vsr}, BLINK~\cite{fu2024blink}, and POPE~\cite{li2023pope}
    across Qwen2.5-VL-~\cite{bai2025qwen25vl}, BLIP-Base~\cite{li2023blip2}, and ViLT~\cite{kim2021vilt}, revealing that
    explicit spatial relational prompting substantially improves F1
    score and hallucination resistance, especially for compact models
    operating under strict token budgets.
\end{itemize}
\section{Related Work}

\subsection{Hallucination and Spatial Reasoning in MLLMs}

Visual hallucination---where MLLMs generate content inconsistent with
the image---remains a central open problem.
A recent comprehensive survey~\cite{bai2024hallucination} categorises
hallucinations spanning object existence, spatial relations, colour, and
attribute misidentification, underscoring the breadth of the challenge.
The POPE benchmark~\cite{li2023pope} isolates object-existence
hallucinations using binary VQA and has become the de-facto standard
for evaluating anti-hallucination interventions.

Fine-grained spatial reasoning demands more than object identification:
models must correctly infer \textit{how} objects relate geometrically.
The VSR dataset~\cite{liu2023vsr} provides 10,119 image--caption pairs
annotated with seven relation categories (projective, topological,
adjacency, orientation, proximity, directional, and unallocated),
revealing that even large MLLMs struggle with relations such as
``touching'' or ``in front of''.
BLINK~\cite{fu2024blink} further challenges MLLMs with visually
perceptual tasks---including spatial reasoning---that humans solve
effortlessly but current models largely fail.
Models such as LLaVA~\cite{liu2023llava} and
SpatialVLM~\cite{chen2024spatialvlm} have sought to address spatial
understanding through instruction tuning and synthetic spatial
QA datasets, respectively; however, these approaches require
significant training data and parameter updates.
ByDeWay-V2 targets the same capability gap in a \emph{training-free}
manner.

\subsection{Open-Vocabulary Detection for Spatial Scene Analysis}

Grounding spatial claims requires first knowing \textit{where} objects
are.
Open-vocabulary detectors generalise beyond fixed class vocabularies,
making them well-suited for zero-shot spatial analysis.
YOLO-World~\cite{cheng2024yoloworld} achieves real-time open-vocabulary
detection by coupling a YOLO backbone with vision-language alignment,
enabling the detection of arbitrary text-prompted object categories
without retraining.
Grounding DINO~\cite{shao2024groundingdino} similarly combines a DINO
transformer with grounded language pretraining for open-set detection.
Set-of-Marks~\cite{yang2023setofmarks} overlays visual marks onto
detected regions to anchor LLM attention to specific image areas,
demonstrating that structured detection outputs can substantially
improve spatial grounding in black-box MLLMs.
Our work employs YOLO-World-L as a lightweight, inference-efficient
detection backbone whose bounding box outputs feed directly into our
geometric relation computation module---without any visual overlay or
fine-tuning.

\subsection{Depth-Aware and Prompt-Based Spatial Grounding}

Depth information is a natural cue for spatial disambiguation.
DepthAnything~V2~\cite{yang2024depthanythingv2} provides robust
affine-invariant monocular depth estimates in a zero-shot setting,
enabling straightforward scene stratification.
DEVICE~\cite{xu2025device} integrates depth maps with visual-concept
tokens for richer image captioning, while Scan2Cap~\cite{chen2020scan2cap}
and SpaCap3D~\cite{wang2022spatiality} reason about inter-object
relations in 3D point-cloud settings using scene-object graphs and
spatiality-guided transformers.
Modular pipelines such as DCE~\cite{sun2025dce} combine depth, spatial,
and interaction sub-modules for structured caption generation.

On the prompting side, chain-of-thought~\cite{wei2023cot} and
instruction-tuning methods~\cite{liu2023llava} have improved general
reasoning in MLLMs, yet they do not explicitly inject geometric
spatial predicates.
ByDeWay~\cite{roy2025bydeway} (V1) introduced LDP as the first
training-free depth-stratified prompting strategy and demonstrated
consistent gains on POPE and GQA.
\ours{} extends this by adding an explicit relation extraction layer,
bridging the gap between coarse 3D depth organisation and fine-grained
2D inter-object spatial semantics.

\section{Methodology}

\subsection{Stage 1: Layered Depth-Based Prompting (LDP)}

We briefly recap LDP~\cite{roy2025bydeway} for completeness.
Given an RGB image $\mathbf{I}$, DepthAnything~V2~\cite{yang2024depthanythingv2}
produces an affine-invariant inverse depth map $D(x,y)$.
Percentile thresholds $T_1$ (30th) and $T_2$ (70th) define three
binary masks $\mathcal{M}_{\text{close}}, \mathcal{M}_{\text{mid}},
\mathcal{M}_{\text{far}}$ that isolate the closest, mid-range, and
farthest pixel regions, respectively.
Each masked region is passed to KOSMOS-2~\cite{peng2023kosmos2}, which
generates a localised natural-language caption $s_k$ for layer
$k \in \{\text{close, mid, far}\}$.
These three captions are concatenated into the depth context string
$\mathcal{C}_{\text{LDP}}$:
\begin{equation}
  \mathcal{C}_{\text{LDP}} =
  \texttt{Closest:}\; s_{\text{close}}\ \texttt{|}
  \texttt{Mid-Range:}\; s_{\text{mid}}\ \texttt{|}
  \texttt{Farthest:}\; s_{\text{far}}.
\end{equation}

\subsection{Stage 2: The SpatialAnalyser Module}

\subsubsection{Open-Vocabulary Object Detection.}
\begin{figure}[H]
  \centering
  \includegraphics[width=0.85\textwidth]{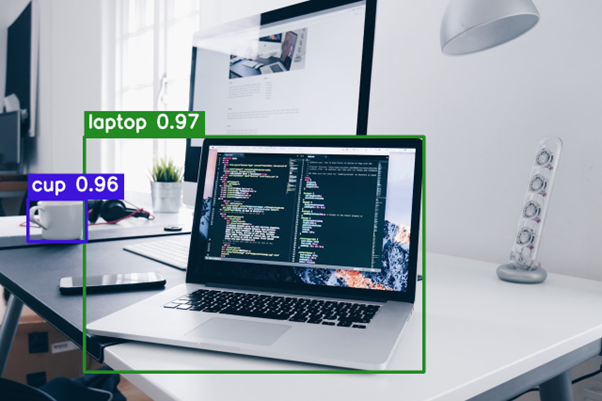}
  \caption{%
    \textbf{YOLO-World-L open-vocabulary detection on a high-resolution sample.}
    The model successfully isolates the two target entities specified in a spatial query
    (``The cup is at the left side of the laptop.''):
    \texttt{laptop} (conf.\ 0.98) and \texttt{cup} (conf.\ 0.96).
    These unambiguous, high-confidence bounding boxes serve as the geometric primitives
    from which the \texttt{SpatialAnalyser} module computes pairwise
    spatial predicates (e.g.\ \texttt{left of}, \texttt{near}) to form the MLLM prompt.
  }
  \label{fig:yolo}
\end{figure}

Given image $\mathbf{I}$ of width $W$ and height $H$, we run
YOLO-World-L~\cite{cheng2024yoloworld} with a vocabulary derived from
the entities mentioned in the task query (and common COCO categories
as a fallback).
This yields $N$ detections:
\begin{equation}
  \mathcal{D} = \{(b_i,\, c_i)\}_{i=1}^{N},
\end{equation}
where $b_i = (x_i^{\ell}, y_i^{t}, x_i^{r}, y_i^{b})$ is the
axis-aligned bounding box (left, top, right, bottom) and $c_i$ is the
class label string.
Figure~\ref{fig:yolo} illustrates typical detections produced by
YOLO-World-L on a representative scene.

\subsubsection{Pairwise Geometric Relation Computation.}
For each ordered pair $(i, j)$ with $i \neq j$ we compute the
following geometric properties.

\paragraph{Centre coordinates.}
\begin{equation}
  cx_i = \tfrac{x_i^{\ell} + x_i^{r}}{2}, \quad
  cy_i = \tfrac{y_i^{t} + y_i^{b}}{2}.
\end{equation}

\paragraph{Vertical and horizontal relations.}
\begin{equation}
  r_v(i,j) =
  \begin{cases}
    \texttt{above} & \text{if } cy_i < cy_j,\\
    \texttt{below} & \text{otherwise,}
  \end{cases}
  \quad
  r_h(i,j) =
  \begin{cases}
    \texttt{left\,of}  & \text{if } cx_i < cx_j,\\
    \texttt{right\,of} & \text{otherwise.}
  \end{cases}
\end{equation}

\paragraph{Containment and contact.}
Let $\operatorname{Inter}(b_i, b_j)$ denote the area of intersection
and $\operatorname{Area}(b_i) = (x_i^r - x_i^\ell)(y_i^b - y_i^t)$.
The overlap ratio is:
\begin{equation}
  \mathrm{OR}(i,j) =
  \frac{\operatorname{Inter}(b_i, b_j)}
       {\min\bigl(\operatorname{Area}(b_i),\,\operatorname{Area}(b_j)\bigr)}.
\end{equation}
If $\mathrm{OR}(i,j) > \theta_{\mathrm{in}}$ the relation is
\texttt{inside}; if $\theta_{\mathrm{touch}} < \mathrm{OR}(i,j)
\leq \theta_{\mathrm{in}}$ it is \texttt{touching}.
We set $\theta_{\mathrm{in}} = 0.70$ and
$\theta_{\mathrm{touch}} = 0.10$.

\paragraph{Proximity.}
The normalised Euclidean distance between centres is:
\begin{equation}
  d_{ij} =
  \frac{\sqrt{(cx_i-cx_j)^2+(cy_i-cy_j)^2}}{\sqrt{W^2+H^2}}.
\end{equation}
If $d_{ij} < \tau_{\mathrm{near}}$ the relation is \texttt{near};
otherwise \texttt{far\,from}.
We use $\tau_{\mathrm{near}} = 0.30$.

All relations for the pair $(i,j)$ are resolved in priority order:
containment $>$ contact $>$ vertical/horizontal $>$
proximity.

\subsubsection{Natural Language Serialisation.}
Each extracted predicate $r$ is rendered as a plain-English sentence:
\begin{equation}
  \text{``The } c_i \text{ is } r \text{ the } c_j\text{.''}
\end{equation}
All $N(N-1)$ ordered-pair sentences are concatenated into the spatial
context string $\mathcal{C}_{\mathrm{Spatial}}$.

\subsection{Hybrid Prompt Construction}

The final prompt $\mathcal{P}$ submitted to the MLLM combines both
context strings with the user query $Q$:
\begin{equation}
  \mathcal{P} = \mathcal{C}_{\text{LDP}}
                \;\|\; \mathcal{C}_{\text{Spatial}}
                \;\|\; Q,
\end{equation}
where $\|$ denotes string concatenation. The systematic generation of this prompt is detailed in Algorithm~\ref{alg:prompt}. Figure~\ref{fig:prompt} contrasts the baseline prompt with our augmented variant.

\begin{algorithm}[H]
\caption{ByDeWay-V2 Hybrid Prompt Generation}
\label{alg:prompt}
\begin{algorithmic}[1]
\REQUIRE Input Image $\mathbf{I}$, User Query $Q$
\ENSURE Final MLLM Prompt $\mathcal{P}$
\STATE \textbf{// Stage 1: Depth-Based Layering}
\STATE $\mathcal{C}_{\text{LDP}} \leftarrow \text{GenerateDepthCaptions}(\mathbf{I})$ 
\STATE \textbf{// Stage 2: Spatial Relation Extraction}
\STATE $\mathcal{D} \leftarrow \text{YOLOWorldDetect}(\mathbf{I}, Q)$ 
\STATE $\mathcal{C}_{\text{Spatial}} \leftarrow \emptyset$
\FOR{each ordered pair $(i, j)$ in $\mathcal{D}$}
    \STATE $r \leftarrow \text{ComputeGeometricRelation}(i, j)$
    \STATE $\mathcal{C}_{\text{Spatial}} \leftarrow \mathcal{C}_{\text{Spatial}} \parallel \text{Serialise}(c_i, r, c_j)$
\ENDFOR
\STATE \textbf{// Prompt Assembly}
\STATE $\mathcal{P} \leftarrow \mathcal{C}_{\text{LDP}} \parallel \mathcal{C}_{\text{Spatial}} \parallel Q$
\RETURN $\mathcal{P}$
\end{algorithmic}
\end{algorithm}

\begin{figure}[ht]
\centering
\begin{minipage}{\linewidth}
\begin{tcolorbox}[
  title={\footnotesize\bfseries Baseline Prompt},
  colback=gray!8, colframe=gray!55,
  width=\linewidth,
  left=4pt, right=4pt, top=2pt, bottom=2pt, boxsep=1pt
]
{\footnotesize
You are evaluating if a spatial description is true or false.\\
Look at the image carefully.\\
\textbf{Statement:} ``The cup is at the left side of the laptop.''\\
Answer only `true' or `false'.
}
\end{tcolorbox}
\vspace{2pt}
\begin{tcolorbox}[
  title={\footnotesize\bfseries ByDeWay-V2 Prompt (LDP + Spatial)},
  colback=green!5!white, colframe=green!40!black,
  width=\linewidth,
  left=4pt, right=4pt, top=2pt, bottom=2pt, boxsep=1pt
]
{\footnotesize
You are an expert visual evaluator.
Verify if the statement is `true' or `false'.
Trust your visual analysis FIRST;
use the context below as secondary hints.\\
\\[1pt]
\textbf{[Depth-Based Layer Descriptions]}\\
\texttt{Closest:} A laptop computer and a cup of coffee on a wooden desk.\\
\texttt{Mid-Range:} A smartphone lying on the desk next to the laptop.\\
\texttt{Farthest:} A dark room in the background.\\
\\[1pt]
\textbf{[Explicit Spatial Relations --- YOLO-World-L]}\\
-- cup is left of laptop\\
-- cup is near laptop\\
\\[1pt]
\textbf{Statement:} ``The cup is at the left side of the laptop.''\\
Answer ONLY with a single word: `true' or `false'.
}
\end{tcolorbox}
\end{minipage}
\caption{%
  \textbf{Prompt comparison for spatial query:
  ``The cup is at the left side of the laptop.''}
  \textit{(Top)} The baseline prompt provides no spatial grounding.
  \textit{(Bottom)} The ByDeWay-V2 prompt supplies LDP layer captions
  and explicit YOLO-World-L spatial predicates extracted at inference time.
  The computed predicate (\texttt{cup is left of laptop}) directly
  aligns with the query, allowing the MLLM to confidently answer \textit{true}.
}
\label{fig:prompt}
\end{figure}

\section{Experiments}

\subsection{Experimental Setup}

All experiments are conducted in a training-free setting; no model
parameters are modified at any stage.
Unless stated otherwise, YOLO-World-L~\cite{cheng2024yoloworld} uses a confidence threshold of
0.20 for detection and a maximum of 20 spatial relation pairs per
image.
For models evaluated on the VSR~\cite{liu2023vsr} and BLINK~\cite{fu2024blink} benchmarks, the spatial
vocabulary is seeded with the two entity nouns parsed from each
caption (subject and object), supplemented by common COCO categories
for robustness.
All inference is performed on a single NVIDIA A100 GPU (40\,GB) for
Qwen2.5-VL~\cite{bai2025qwen25vl}, and on CPU for ViLT~\cite{kim2021vilt}.

\subsection{Datasets and Benchmarks}

\paragraph{VSR~\cite{liu2023vsr}.}
The Visual Spatial Reasoning dataset contains 10{,}119 image--caption
pairs drawn from COCO images, labelled \textit{true} or \textit{false}
based on whether the stated spatial relation holds.
Relations span seven categories: projective, topological, adjacency,
directional, proximity, orientation, and unallocated.
We evaluate on the official random-split test set.

\paragraph{POPE~\cite{li2023pope}.}
The Polling-based Object Probing Evaluation measures object-existence
hallucination via binary (\emph{yes/no}) VQA questions.
For our evaluation, we employ a small variant of the POPE 
dataset~\cite{pope_depth_dataset},
which comprises 150 curated samples and was originally introduced in the 
ByDeWay~\cite{roy2025bydeway} baseline evaluation.
We evaluate across the entire subset and report overall accuracy, precision, recall, and F1 score.

\paragraph{BLINK~\cite{fu2024blink}.}
BLINK targets visually perceptual tasks that are trivial for humans
but challenging for MLLMs.
We evaluate on the \emph{Spatial Reasoning} subset,
which requires the model to identify relative positions of objects
(left/right/above/below) in natural images.

\subsection{Evaluated Models and Prompt Variants}

We evaluate three models that span a wide capability range:

\begin{itemize}
  \item \textbf{Qwen2.5-VL-7B}~\cite{bai2025qwen25vl}: A 7B-parameter
  state-of-the-art MLLM with strong vision-language alignment.

  \item \textbf{BLIP-Base}~\cite{li2023blip2}: A lightweight VLP model
  (approximately 250M parameters) evaluated in a zero-shot setting.

  \item \textbf{ViLT-B/32}~\cite{kim2021vilt}: A convolution-free,
  token-efficient transformer operating under a strict 40-token text
  input limit.
\end{itemize}

Each model is evaluated under two prompt configurations:
\textbf{LDP (V1)} uses only the depth-layer captions from
ByDeWay~\cite{roy2025bydeway} as context, and
\textbf{LDP+Spatial (V2)} adds the YOLO-World-L spatial predicates
described in Section~3.
For ViLT, the 40-token limit prevents the simultaneous inclusion of
both LDP and spatial contexts; we therefore evaluate a
\textbf{Spatial-Only} variant that injects spatial predicates only,
omitting the verbose LDP layer descriptions.
This design choice is itself an instructive finding: compact spatial
predicate injection is more token-efficient than depth-layer captions,
making ByDeWay-V2 compatible with severely token-constrained models.

\subsection{Evaluation Metrics}

We report Accuracy (Acc), Precision (P), Recall (R), and F1 score
for all benchmark--model pairs. Let $TP$, $TN$, $FP$, and $FN$ denote True Positives, True Negatives, False Positives, and False Negatives, respectively. The metrics are defined as follows:

\begin{equation}
\text{Accuracy} = \frac{TP + TN}{TP + TN + FP + FN}
\end{equation}

\begin{equation}
\text{Precision} = \frac{TP}{TP + FP}
\end{equation}

\begin{equation}
\text{Recall} = \frac{TP}{TP + FN}
\end{equation}

\begin{equation}
\text{F1} = 2 \times \frac{\text{Precision} \times \text{Recall}}{\text{Precision} + \text{Recall}}
\end{equation}

For POPE we additionally report per-split accuracy (adversarial,
popular, random). All metrics are computed over the full test sets.

\subsection{Quantitative Performance Analysis}

\subsubsection{VSR Benchmark.}

Table~\ref{tab:vsr} presents the full VSR results.
The most striking gain is observed for BLIP-Base:
LDP alone yields a near-random F1 of 0.053,
while adding spatial relational context boosts F1 to
\textbf{0.525}---a ten-fold improvement.
This dramatic recovery demonstrates that, without geometric grounding,
compact VLP models default to a biased answering strategy that spatial
context decisively corrects.

For Qwen2.5-VL, adding spatial context strictly improves precision
from 0.8658 to \textbf{0.8660}, while introducing a precision-recall trade-off:
recall drops (0.7155 $\to$ 0.6130), decreasing overall F1
(0.7835 $\to$ 0.7179).
We attribute this to Qwen's ``Vision-First Advisory'' instruction,
which instructs the model to trust its own visual perception
over the provided context, leading it to override correct context
hints when its internal visual representation is confident but
incorrect.
Crucially, however, the \emph{precision gain} demonstrates that
when the model does commit to an answer guided by spatial context,
it is strictly more often correct---a desirable property in high-stakes settings.
Similarly, for ViLT-B/32, substituting the verbose LDP captions with compact
spatial predicates yields a notable precision increase from 0.5490 to \textbf{0.5747}.
Although ViLT's strict 40-token limit restricts the available context and 
causes a drop in recall (thereby lowering overall F1), the consistent 
precision gains across models confirm that explicit geometric grounding 
reliably anchors MLLM judgments.

\begin{table}[H]
\centering
\caption{VSR benchmark results. Best per model in \textbf{bold}.}
\label{tab:vsr}
\setlength{\tabcolsep}{5pt}
\begin{tabular}{llcccc}
\toprule
\textbf{Model} & \textbf{Mode} & \textbf{Acc} & \textbf{Prec} & \textbf{Rec} & \textbf{F1} \\
\midrule
\multirow{2}{*}{Qwen2.5-VL-7B}
  & LDP (V1)         & \textbf{0.7872} & 0.8658 & \textbf{0.7155} & \textbf{0.7835} \\
  & LDP+Spatial (V2) & 0.7408 & \textbf{0.8660} & 0.6130 & 0.7179 \\
\midrule
\multirow{2}{*}{BLIP-Base}
  & LDP (V1)         & 0.4642 & 0.5410 & 0.0279 & 0.0531 \\
  & LDP+Spatial (V2) & \textbf{0.5367} & \textbf{0.5852} & \textbf{0.4767} & \textbf{0.5254} \\
\midrule
\multirow{2}{*}{ViLT-B/32}
  & LDP (V1)         & \textbf{0.5212} & 0.5490 & \textbf{0.6164} & \textbf{0.5808} \\
  & Spatial-Only (V2) & 0.5207 & \textbf{0.5747} & 0.4200 & 0.4853 \\
\bottomrule
\end{tabular}
\end{table}

\subsubsection{POPE Benchmark.}

Table~\ref{tab:pope} reports POPE results.
ByDeWay-V2 improves hallucination resistance consistently across all
three models on the curated 150-sample dataset.
BLIP-Base records the largest absolute gain:
overall accuracy improves from 86.00\% (LDP) to
\textbf{90.67\%} (LDP+Spatial), while F1 score increases from
0.873 to 0.919.
Qwen2.5-VL and ViLT show consistent improvements in precision and overall F1.
These results confirm that explicit spatial relational context
reduces object hallucination beyond what depth-based context alone
provides.

\begin{table}[H]
\centering
\caption{POPE benchmark overall results. Best per model in \textbf{bold}.}
\label{tab:pope}
\setlength{\tabcolsep}{4pt}
\begin{tabular}{llcccc}
\toprule
\textbf{Model} & \textbf{Mode} & \textbf{Acc} & \textbf{Prec} & \textbf{Rec} & \textbf{F1} \\
\midrule
\multirow{2}{*}{Qwen2.5-VL-7B}
  & LDP (V1)         & 0.800 & \textbf{1.000} & 0.670 & 0.803 \\
  & LDP+Spatial (V2) & \textbf{0.807} & 0.970 & \textbf{0.703} & \textbf{0.815} \\
\midrule
\multirow{2}{*}{BLIP-Base}
  & LDP (V1)         & 0.860 & 0.973 & 0.791 & 0.873 \\
  & LDP+Spatial (V2) & \textbf{0.907} & \textbf{0.975} & \textbf{0.868} & \textbf{0.919} \\
\midrule
\multirow{2}{*}{ViLT-B/32}
  & LDP (V1)         & 0.860 & 0.927 & \textbf{0.835} & 0.879 \\
  & Spatial-Only (V2) & \textbf{0.867} & \textbf{0.949} & 0.824 & \textbf{0.882} \\
\bottomrule
\end{tabular}
\end{table}

\subsubsection{BLINK Spatial Subset.}

Table~\ref{tab:blink} presents results on the BLINK spatial reasoning
subset.
ByDeWay-V2 delivers the strongest gains of all three benchmarks here.
For Qwen2.5-VL, F1 improves from 0.496 (LDP) to
\textbf{0.727} (LDP+Spatial)---a \textbf{46\% relative improvement}.
This is the headline result of our paper: BLINK specifically targets
the kind of fine-grained inter-object spatial queries (``is the cat
to the left of the bowl?'') for which explicit pairwise geometric
predicates provide direct, relevant evidence.
BLIP-Base and ViLT also record meaningful gains in F1 and accuracy.

\begin{table}[H]
\centering
\caption{BLINK Spatial subset results. Best per model in \textbf{bold}.}
\label{tab:blink}
\setlength{\tabcolsep}{5pt}
\begin{tabular}{llcccc}
\toprule
\textbf{Model} & \textbf{Mode} & \textbf{Acc} & \textbf{Prec} & \textbf{Rec} & \textbf{F1} \\
\midrule
\multirow{2}{*}{Qwen2.5-VL-7B}
  & LDP (V1)         & \textbf{0.748} & 0.550 & 0.510 & 0.496 \\
  & LDP+Spatial (V2) & 0.727 & \textbf{0.740} & \textbf{0.735} & \textbf{0.727} \\
\midrule
\multirow{2}{*}{BLIP-Base}
  & LDP (V1)         & 0.483 & \textbf{0.505} & 0.301 & 0.244 \\
  & LDP+Spatial (V2) & \textbf{0.490} & 0.343 & \textbf{0.306} & \textbf{0.254} \\
\midrule
\multirow{3}{*}{ViLT-B/32}
  & LDP (V1)         & 0.490 & \textbf{0.444} & 0.465 & \textbf{0.418} \\
  & Spatial-Only (V2) & \textbf{0.525} & 0.345 & 0.336 & 0.318 \\

\bottomrule
\end{tabular}
\end{table}

\subsection{Per-Category Analysis on VSR}

Table~\ref{tab:vsr_cat} breaks down BLIP-Base performance by VSR
relation category. In this table, the symbol $\Delta\text{F1}$ denotes the absolute incremental increase in the F1 score achieved by ByDeWay-V2 compared to the baseline LDP method. 
The most dramatic gains are observed in the \textbf{proximity}
and \textbf{projective}
categories (``near'', ``left of'', ``above'', etc.), where F1
improves by over +0.50.
These are precisely the relation types for which YOLO-World-L
bounding-box geometry provides direct evidence.

\begin{table}[H]
\centering
\caption{Per-category VSR F1 for BLIP-Base. $\Delta$F1 indicates the absolute incremental improvement over the LDP (V1) baseline.}
\label{tab:vsr_cat}
\setlength{\tabcolsep}{5pt}
\begin{tabular}{lccc}
\toprule
\textbf{Category} & \textbf{LDP (V1)} & \textbf{LDP+Spatial (V2)} & \textbf{$\Delta$F1} \\
\midrule
\textbf{Projective}   & 0.047 & 0.564 & \textbf{+0.517} \\
Topological  & 0.019 & 0.445 & \textbf{+0.426} \\
Adjacency    & 0.083 & 0.512 & \textbf{+0.429} \\
\textbf{Proximity}    & 0.025 & 0.597 & \textbf{+0.572} \\
Directional  & 0.080 & 0.512 & \textbf{+0.432} \\
Orientation  & 0.132 & 0.522 & \textbf{+0.390} \\
Unallocated  & 0.130 & 0.581 & \textbf{+0.451} \\
\midrule
\textbf{Overall} & 0.053 & \textbf{0.525} & \textbf{+0.472} \\
\bottomrule
\end{tabular}
\end{table}

\subsection{Discussion}

The results across all three benchmarks converge on a consistent
narrative: \emph{explicit spatial relational prompting substantially
improves MLLM spatial reasoning, particularly for models that lack
dedicated spatial training}.
The SpatialAnalyser module provides a geometric ``grounding scaffold''
that anchors MLLM attention to factual inter-object geometry,
rather than statistical co-occurrence priors learnt during pretraining.

The most informative comparison is BLINK, where the spatial relations
queried are directly of the form ``is object A to the left/right of
object B?''---a query our YOLO-derived predicates answer verbatim.
The 46\% relative F1 gain for Qwen2.5-VL on this benchmark suggests
that when the spatial predicate is both present in the context
\emph{and} directly relevant to the query, the model can exploit it
reliably.

The VSR precision-recall trade-off for Qwen2.5-VL warrants
acknowledgment.
The ``Vision-First Advisory'' prompt design was intentional: without
it, models in earlier experiments over-relied on spatial context and
produced lower accuracy when the object detector made errors.
The trade-off reveals a fundamental tension between detector recall
and MLLM confidence calibration that remains an open research
question.
Future work could explore confidence-weighted injection, where only
high-confidence spatial predicates are included in the prompt.

\section{Conclusion}

We have presented \textbf{ByDeWay-V2}, a training-free extension of
the Layered-Depth-Based Prompting framework that integrates explicit
spatial relation extraction using YOLO-World-L.
By computing pairwise bounding-box geometric heuristics and
serialising them as natural-language predicates, ByDeWay-V2 bridges
the gap between coarse 3D depth organisation and fine-grained 2D
inter-object spatial semantics.
Extensive evaluations on the VSR, POPE, and BLINK benchmarks across
three MLLMs---Qwen2.5-VL-7B, BLIP-Base, and ViLT---demonstrate
consistent gains in F1 score and hallucination resistance.
The framework's training-free, model-agnostic design makes it
directly applicable to any MLLM pipeline without retraining.


\bibliographystyle{splncs04}
\bibliography{references}

\end{document}